\documentclass[journal,twocolumn]{IEEEtran}

\usepackage[mathscr]{eucal}
\usepackage[cmex10]{amsmath}
\usepackage{epsfig,epsf,psfrag}
\usepackage{amssymb,amsmath,amsthm,amsfonts,latexsym}
\usepackage{graphicx}
\usepackage{amsmath,bm,xcolor,url}
\usepackage[caption=false]{subfig} 
\usepackage{array}
\usepackage{verbatim}
\usepackage{bm}
\usepackage{algorithmic}
\usepackage{algorithm}
\usepackage{verbatim}
\usepackage{textcomp}
\usepackage{mathrsfs}
\usepackage{epstopdf}
\usepackage{cite}

\catcode`~=11 \def\UrlSpecials{\do\~{\kern -.15em\lower .7ex\hbox{~}\kern .04em}} \catcode`~=13 

\allowdisplaybreaks[3]

\newcommand{\calC}{\mathcal{C}}

\newcommand{\calF}{\mathcal{F}}

\newcommand{\calI}{\mathcal{I}}

\newcommand{\calR}{\mathcal{R}}

\newcommand{\ba}{\mathbf{a}}
\newcommand{\bA}{\mathbf{A}}

\newcommand{\bB}{\mathbf{B}}

\newcommand{\bC}{\mathbf{C}}

\newcommand{\be}{\mathbf{e}}

\newcommand{\bF}{\mathbf{F}}
\newcommand{\bg}{\mathbf{g}}
\newcommand{\bG}{\mathbf{G}}

\newcommand{\bH}{\mathbf{H}}

\newcommand{\br}{\mathbf{r}}

\newcommand{\bS}{\mathbf{S}}

\newcommand{\bu}{\mathbf{u}}
\newcommand{\bU}{\mathbf{U}}
\newcommand{\bv}{\mathbf{v}}
\newcommand{\bV}{\mathbf{V}}
\newcommand{\bw}{\mathbf{w}}

\newcommand{\bx}{\mathbf{x}}

\newcommand{\by}{\mathbf{y}}

\newcommand{\bz}{\mathbf{z}}

\newcommand{\bbR}{\mathbb{R}}

\DeclareMathAlphabet{\mathbsf}{OT1}{cmss}{bx}{n}
\DeclareMathAlphabet{\mathssf}{OT1}{cmss}{m}{sl}

\DeclareSymbolFont{bsfletters}{OT1}{cmss}{bx}{n}  
\DeclareSymbolFont{ssfletters}{OT1}{cmss}{m}{n}
\DeclareMathSymbol{\bsfGamma}{0}{bsfletters}{'000}
\DeclareMathSymbol{\ssfGamma}{0}{ssfletters}{'000}
\DeclareMathSymbol{\bsfDelta}{0}{bsfletters}{'001}
\DeclareMathSymbol{\ssfDelta}{0}{ssfletters}{'001}
\DeclareMathSymbol{\bsfTheta}{0}{bsfletters}{'002}
\DeclareMathSymbol{\ssfTheta}{0}{ssfletters}{'002}
\DeclareMathSymbol{\bsfLambda}{0}{bsfletters}{'003}
\DeclareMathSymbol{\ssfLambda}{0}{ssfletters}{'003}
\DeclareMathSymbol{\bsfXi}{0}{bsfletters}{'004}
\DeclareMathSymbol{\ssfXi}{0}{ssfletters}{'004}
\DeclareMathSymbol{\bsfPi}{0}{bsfletters}{'005}
\DeclareMathSymbol{\ssfPi}{0}{ssfletters}{'005}
\DeclareMathSymbol{\bsfSigma}{0}{bsfletters}{'006}
\DeclareMathSymbol{\ssfSigma}{0}{ssfletters}{'006}
\DeclareMathSymbol{\bsfUpsilon}{0}{bsfletters}{'007}
\DeclareMathSymbol{\ssfUpsilon}{0}{ssfletters}{'007}
\DeclareMathSymbol{\bsfPhi}{0}{bsfletters}{'010}
\DeclareMathSymbol{\ssfPhi}{0}{ssfletters}{'010}
\DeclareMathSymbol{\bsfPsi}{0}{bsfletters}{'011}
\DeclareMathSymbol{\ssfPsi}{0}{ssfletters}{'011}
\DeclareMathSymbol{\bsfOmega}{0}{bsfletters}{'012}
\DeclareMathSymbol{\ssfOmega}{0}{ssfletters}{'012}

\newcommand{\btheta}{\bm{\theta}}
\newcommand{\btau}{\bm{\tau}}

\newcommand{\brho}{\bm{\rho}}

\newcommand{\bPsi}{\bm{\Psi}}

\newcommand{\bPhi}{\bm{\Phi}}

\DeclareMathOperator*{\argmin}{arg\,min}

\DeclareMathOperator{\diag}{diag}

\DeclareMathOperator{\vect}{vec}

\newcommand{\bzero}{\mathbf{0}}
\newcommand{\bone}{\mathbf{1}}

\newcommand{\qednew}{\nobreak \ifvmode \relax \else
      \ifdim\lastskip<1.5em \hskip-\lastskip
      \hskip1.5em plus0em minus0.5em \fi \nobreak
      \vrule height0.75em width0.5em depth0.25em\fi}

\usepackage{multirow}
\usepackage{booktabs}

\newcommand{\abs}[1] {\left| {#1} \right|}

\newcommand{\micron}{{\mu\text{m}}}

\newcommand{\dotprod} {\cdot}

\newcommand{\hadprod} {{\odot}}
\DeclareMathOperator{\diagmat}{Diag}
\newcommand{\norm}[1] {\left\Vert{#1}\right\Vert}

\DeclareMathOperator{\re}{\mathbb{R}e}
\DeclareMathOperator{\im}{\mathbb{I}m}
\def \realpart#1 {{\re \! \left[ #1 \right]}}
\def \imagpart#1 {{\im \! \left[ #1 \right]}}

\newcommand{\measurevec} {\by}
\newcommand{\ambientcoeff} {\alpha}
\newcommand{\ambientvec} {\ba}
\newcommand{\Coh} {W}
\newcommand{\CohUnrotated} {W}

\newcommand{\sparsepropmat} {\bU}

\newcommand{\lagrangemult} {\by}
\newcommand{\SI} {{\hat{I}}}
\newcommand{\Intensity} {{I}}

\newcommand{\cohwidth} {{\sigma}}
\newcommand{\penaltyweight} {\beta}
\newcommand{\sparseweight} {\kappa}
\newcommand{\measureweight} {\mu}
\newcommand{\sparsemat}{\bPsi}

\newcommand{\sensemat}{\bPhi}
\newcommand{\sensematI}{\bPhi^{I}}
\newcommand{\sensematC}[1]{\bPhi_{#1}^{C}}

\newcommand{\measuremat}{\by}
\newcommand{\measurematI}{\by^{I}}
\newcommand{\measurematC}[1]{\by^{C}_{#1}}

\newcommand{\measRsetC}{\calR}

\newcommand{\measidxsetC}{\calC}
\newcommand{\measidxsetI}{\calI}

\newcommand{\measscale} {\bB}

\newcommand{\normweight} {v}
\newcommand{\normweightvec} {\bv}
\newcommand{\normweightmat} {\bV}

\newcommand{\measnoise} {n}
\newcommand{\measnoiseI} {n_{I}}
\newcommand{\measnoiseC} {n_{C}}

\newcommand{\wallsd} {w}
\newcommand{\wallsdvec} {\bw}

\newcommand{\fineweightcal} {\eta}

\newcommand{\cohzerorad} {p}

\newcommand{\sensemapspatial} {\tau}
\newcommand{\sensemapspatialvec} {\btau}

\newcommand{\sensemapfreq} {\widehat{\sensemapspatial}}
\newcommand{\sensemapfreqvec} {\widehat{\sensemapspatialvec}}

\newcommand{\sourceimgvec} {\bg}
\newcommand{\sourceimgmat} {\bG}
\newcommand{\sourceimg} {I}

\newcommand{\propinttermA} {C}
\newcommand{\propinttermAmat} {\bC}

\newcommand{\propinttermscatter} {S}
\newcommand{\propinttermscattermat} {\bS}

\newcommand{\propinttermB} {H}
\newcommand{\propinttermBmat} {\bH}

\newcommand{\stopthreshprime} {\epsilon^{\text{pri}}}
\newcommand{\stopthreshdual} {\epsilon^{\text{dual}}}
\newcommand{\gradstopthresh} {\epsilon^{\text{grad}}}

\newcommand{\quadpoint} {q}

\newcommand{\vectbrack}[1] {\vect\left\{#1\right\}}

\begin{document}

\title{Multi-Modal Non-line-of-sight Passive Imaging}

\author{
Andre~Beckus~\IEEEmembership{Student Member,~IEEE,} 
Alexandru~Tamasan, and
George~K.~Atia~\IEEEmembership{Senior Member,~IEEE}
\thanks{
This work was supported in part by the Defense Advanced Research Projects Agency under Contract HR0011-16-C-0029 and in part by the National Science Foundation CAREER Award under Grant CCF-1552497.

A. Beckus and G. K. Atia are with the Department of Electrical and Computer Engineering, University of Central Florida, Orlando, FL 32816 USA.

A. Tamasan is with the Department of Mathematics, University of Central Florida, Orlando, FL 32816 USA.
}
}

\maketitle

\begin{abstract}

We consider the non-line-of-sight (NLOS) imaging of an object using the light reflected off a diffusive wall. The wall scatters incident light such that a lens is no longer useful to form an image. Instead, we exploit the 4D \emph{spatial coherence} function to reconstruct a 2D projection of the obscured object. The approach is completely \emph{passive} in the sense that no control over the light illuminating the object is assumed and is compatible with the partially coherent fields ubiquitous in both the indoor and outdoor environments. We formulate a multi-criteria convex optimization problem for reconstruction, which fuses the reflected field's intensity and spatial coherence information at different scales. Our formulation leverages established optics models of light propagation and scattering and exploits the sparsity common to many images in different bases. We also develop an algorithm based on the alternating direction method of multipliers to efficiently solve the convex program proposed. A means for analyzing the null space of the measurement matrices is provided as well as a means for weighting the contribution of individual measurements to the reconstruction. This paper holds promise to advance passive imaging in the challenging NLOS regimes in which the intensity does not necessarily retain distinguishable features and provides a framework for multi-modal information fusion for efficient scene reconstruction.

\end{abstract}


\IEEEpeerreviewmaketitle

\section{Introduction}

\IEEEPARstart{I}{dentifying} an object from indirect light can provide critical information in many practical applications, e.g., in defense, collision avoidance, or emergency situations. 
Non-line-of-sight (NLOS) imaging is a brand new area of research, where state-of-the-art methods either use controllable coherent sources (lasers) as in \cite{Yoo:91}, or require some active modulation of the field after reflection from the object \cite{katz2012looking}. 
Scenes can be imaged by bouncing laser pulses off the wall and measuring intensity \cite{Klein2016}, or time-of-flight \cite{Velten2012,7559994,Buttafava:15}.
A central theme of these works is the isolation of ballistic photons (direct carriers of information about the object) from the diffusely scattered photons, which have lost the history of their interaction with the object. Inherent to these methods, the source is highly coherent and under the control of the observer.

By contrast, here we assume partially coherent sources (e.g. fluorescent or LED light, as well as sunlight), which are outside the control of the observer. We refer to this problem as \textit{Passive} NLOS imaging.

\begin{figure}[t!]
\centering
\includegraphics[scale=1]{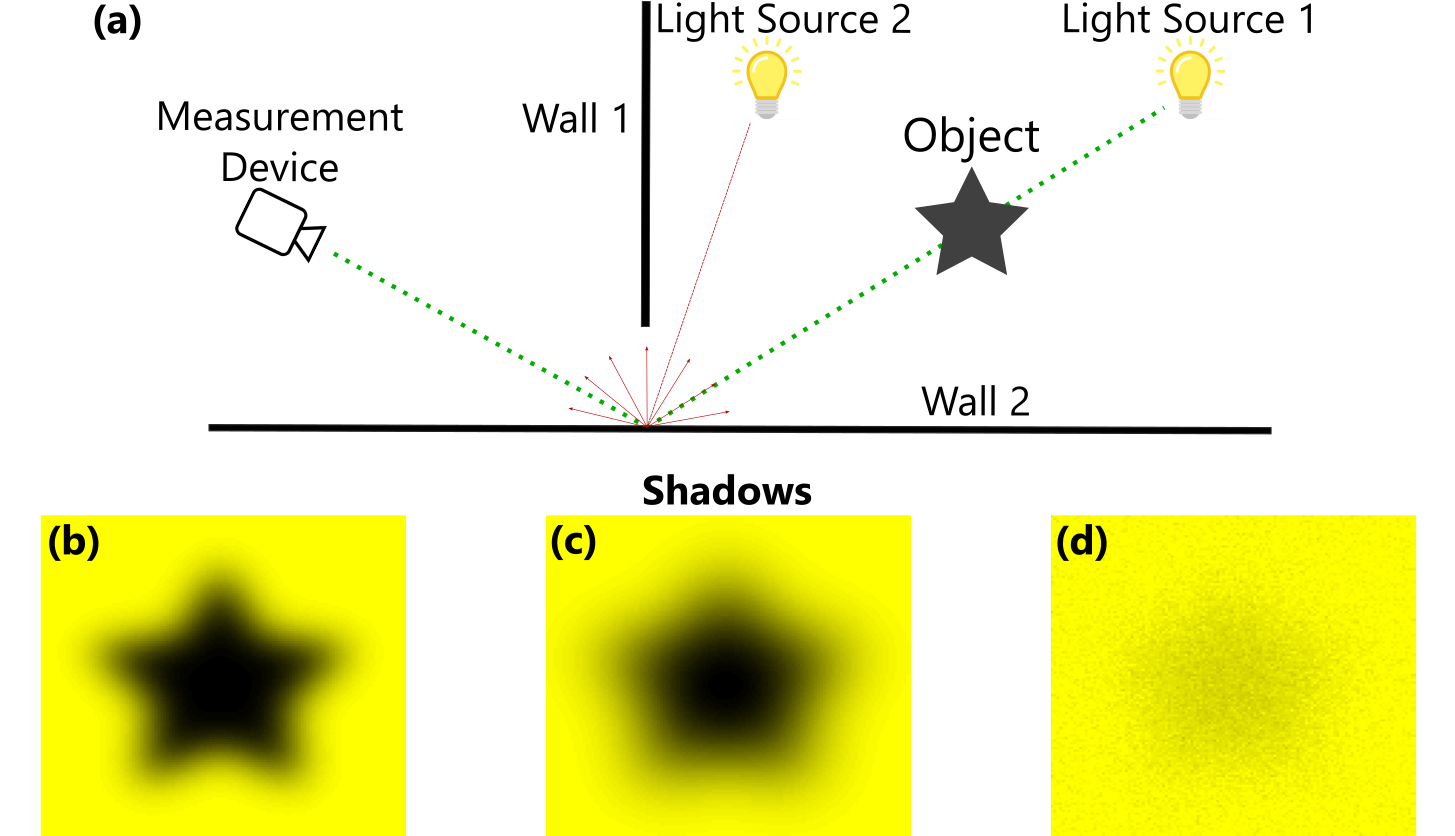}
\caption{
(a) Scenario considered in this paper.
Bottom of figure shows examples of shadows in scenarios ranging from ideal on left to non-ideal on right.
(b) Distinct shadow cast with highly coherent light.
(c) Indistinct shadow due to less coherent light.
(d) Faint and noisy shadow due to Light Source 2 being turned on.  To generate this shadow, uniform ambient light and Gaussian noise are added to (b), and the pixels are then quantized to 16 bits.
}
\label{Fig:Scenario}
\end{figure}

An example of such a problem is illustrated in Fig. \ref{Fig:Scenario}: An object is hidden from view by the obstructing Wall 1, and a complex camera measures light reflected from Wall 2. There may also be a shadow (i.e., spatial variation in intensity pattern) on the wall, whose edge resolution decreases with the decrease of the field's spatial coherence; contrast the highly coherent case in Fig.~\ref{Fig:Scenario}(b) with the lower coherent case in Fig.~\ref{Fig:Scenario}(c).
In addition, a second source floods the wall with light, see Fig.~\ref{Fig:Scenario}(d). While a lensed camera may still be able to image the shadow, the image quality will be degraded due to noise and quantization error.

Existing approaches to the passive imaging problem have relied mostly on intensity-only measurements; for example by assuming some known occlusions are also present, as in the ``accidental'' pinhole camera  \cite{6247698}, or the ``corner'' camera \cite{8237511}.
Other related problems which use intensity or light-field measurements concern imaging through volumetric scattering in turbid media such as fog \cite{7932135,7752955} or water \cite{8237525,Schettini2010}, and their solutions require weak scattering.
However, some recent use of the autocorrelation in intensity at different locations seems promising to work under more relaxed scattering assumptions \cite{Katz2014}.
Phase-space measurements have also been used for imaging \cite{Takasaki:14} or for determining the three-dimensional location of point sources embedded in biological samples \cite{Liu:15}.

The imaging problem of concern here assumes surface scattering is stronger than volumetric scattering.
One such instance was recently demonstrated for scattering from walls at grazing angles \cite{Batarseh2018}.
In ideal cases, this may allow images to be formed from the reflection with a normal lensed camera (recent work even suggests this phenomenon accounts for mirages previously attributed to air temperature differentials \cite{Lu:17}).

Here, we consider the less ideal case, where a useful image cannot be formed using a regular camera, but information is still retained in the spatial coherence of the reflected light.

The spatial coherence $\Coh$ of an electric field
$E$ at two points $\br_1,\br_2$ is defined as an ensemble average over random field realizations $$\Coh(\br_1,\br_2) = \langle E(\br_1) E^*(\br_2) \rangle,$$where $*$ denotes the complex conjugation, and $\langle \cdot \rangle$ is an ensemble average over field realizations (see \cite{Wolf:07}). As customarily used in optics, we work with $\Coh$ in rotated coordinates: the midpoint $\br = (\br_1+\br_2)/{2}$, and displacement $\brho = \br_1-\br_2$, yielding
$\Coh(\br,\brho) = \langle E(\br+\brho/2) E^*(\br-\brho/2) \rangle.$
Note that the intensity of the field $\Intensity(\br) = \Coh(\br, \bzero)$.

In line-of-sight (LOS) imaging, coherence preserves information such that a 3D scene can be reconstructed \cite{Beckus:18}.
Recently, the retention of information was noted experimentally in NLOS \emph{sensing} \cite{Batarseh2018}. Here, we propose an imaging method, which demonstrates the ability to reconstruct discernible 2D projections of obscured objects in NLOS settings, by leveraging the experimentally-verified physical models of \cite{Batarseh2018}.

Coherence has classically been measured using double slits \cite{Thompson:57} with modern experiments realizing the slits using digital micromirror devices \cite{Kondakci17OE}. Many other modern techniques have also emerged including use of shearing interferometers \cite{RezvaniNaraghi:17}, microlens arrays \cite{Stoklasa14NC} and phase-space tomography \cite{Sharma16OE}.

Our approach is physics-driven in the sense that we use established physics-based models from the theory of light propagation and scattering to describe the transformation between the source image and the measurements \cite{7214350}.
The proposed imaging method is based on a multi-modal data fusion.
We formulate and study a convex optimization problem, and propose an algorithm for solving it based on the Alternating Direction Method of Multipliers (ADMM) \cite{Boyd:2011:DOS:2185815.2185816}.
The optimization problem incorporates regularization for sparsity, and reconstructs the image in a suitable transformed basis in which the source image is assumed to have a sparse representation.

In contrast with some existing fusion approaches, which merge multiple \textit{images} in a spatial or wavelet domain \cite{1565303,7707392,4032816}, our method reconstructs a single image by fusing multiple \textit{measurement types} at different spatial scales while exploiting their respective propagation models.
In spirit, our approach to fusion relates to that of \cite{doi:10.1117/1.JMI.4.1.014003}, where a convex optimization problem is devised to pansharpen medical images.

We provide a means of assessing the null space of the model, and a weighting scheme and decision framework by which individual samples of a measurement may be excluded.

The simulated results demonstrate the concept of NLOS imaging using spatial coherence. We further give examples of fusion, and show how the null space of the measurement transformations can be analyzed. 

The paper is organized as follows.  In Section~\ref{section:PhysicalModel}, we review the physical models for propagation and scattering.  In Section~\ref{section:ProposedMethod}, we formulate the NLOS image reconstruction problem and describe the algorithm. The results of running the algorithm are presented in Section~\ref{section:Results}.  In Section~\ref{section:Discussion}, we discuss possible extensions to this work and how our work fits into a practical framework.  The details of the optimization algorithm are described in Appendix~\ref{sec:optalgdetails}, and details of the physical model are given in Appendix~\ref{Sec:CoherenceDerivations}.

\subsection{Notation}

Vectors and matrices are denoted using bold-face lower-case and upper-case letters, respectively.
Given a vector $\ba$, its $\ell_p$-norm is denoted by $\norm{\ba}_p$ and $a(i)$ is the $i\textsuperscript{th}$ element.
The diagonalization operator $\diagmat(\ba)$ returns a matrix with the elements of $\ba$ along the diagonal.
The vectorization of an $M \times N$ matrix $\bA$ is denoted $\vectbrack{\bA}$, with the result taking the form of an $MN$ element vector.
The unit vector with a one in the $i\textsuperscript{th}$ entry is denoted $\be_i$.  Matrices or vectors containing all ones or all zeros are denoted $\bone$ and $\bzero$, respectively, where the dimensions will be clear from the context.
The Hadamard product $\hadprod$ returns the element-wise product of its arguments.
A weighted norm is defined as $\left\| \ba \right\|_{\normweightvec}^2 = \ba^* (\diagmat \normweightvec) \ba$.

The 2D Fourier transform of a function $f(x,y)$ is denoted $\calF \left\{ f(x,y) \right\} (\omega_x, \omega_y)$, where $\omega_x$ and $\omega_y$ are angular frequencies.  The 2D Discrete Fourier Transform (DFT) of matrix $\bA$ is expressed as $\bF_1 \bA \bF_2$, where $\bF_1$ and $\bF_2$ are the 1D DFTs along the columns and rows of $\bA$, respectively. The notation $\star$ is used to indicate both the continuous and discrete forms of the two-dimensional convolution operator.

\section{Physical Model} \label{section:PhysicalModel}

In this section, we describe the physical model.   More details regarding the derivations of the equations can be found in Appendix~\ref{Sec:CoherenceDerivations}. Additional details regarding the models, including experimental verification, can be found in \cite{Batarseh2018}.

We consider the wave model in which a light source emits a random field. It is assumed that light propagates along the longitudinal $z$ direction according to the Fresnel model, where the normals to the wave front make small angles with the direction of propagation.

\begin{figure}[t!]
\centering
\includegraphics[scale=1]{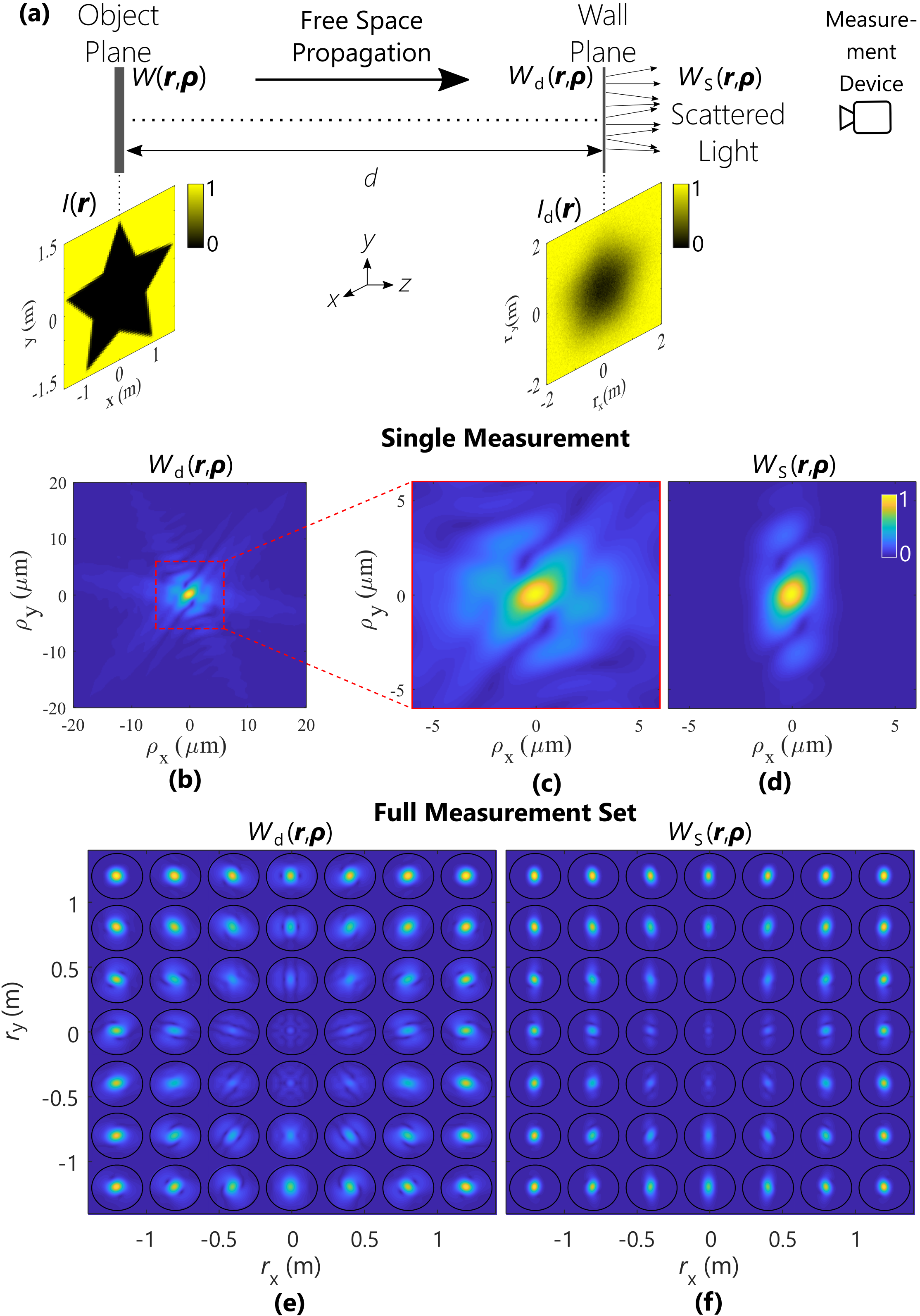}
\vspace{-0.7cm}
\caption{
Details of spatial coherence model. All coherence plots show the magnitude of the coherence function.
(a) Diagram of coherence model, including plots of the intensities in the object plane and wall plane.
(b) Coherence of incident light in wall plane with (c) detail zoom.  Plots are for spatial point $\br=(-0.4~\text{m},-0.4~\text{m})$.
(d) Coherence scattered from wall.
(e) Set of incident coherences plotted on a $7 \times 7$ grid.  Each plot is centered at the corresponding spatial point $\br$.  The radius of each plot is $5.5~\mu$m.
The coherence measurements are shown in the style of light field plots as found for example in \cite{8070368} and \cite{Levoy:2006:LFM:1141911.1141976}.
(f) Scattered coherences as in (d).
}
\label{Fig:FreeSpacePropMeas}
\end{figure}

The approach works with targets in which the projection on the $z$-axis is much smaller compared to the (optical) distance $d$ to the detector (a requirement which is met in many practical situations), thus reducing the problem to that of reconstructing a 2D image; see the illustration in Fig. \ref{Fig:FreeSpacePropMeas}(a).

In the Fresnel model, given a 2D intensity function $\sourceimg(\br)$, the coherence of the light after propagating distance $d$ can be calculated using the linear transformation
\begin{align} \label{eqn:PropagationIntegralCoh}
\Coh_d(\br, \brho)
= \propinttermA(\br,\brho) \, \calF \left\{ \propinttermB(\br-\br') \sourceimg(\br') \right\} \! \left(k \brho / d \right)
\end{align}
where
\begin{align}
\propinttermA(\br,\brho) &=  \frac{2 \pi \cohwidth^2}{\lambda^2 d^2} \exp \left( \frac{i k}{d} \br \dotprod \brho \right),
\label{eqn:PropagationIntegralCohA}
\\
\propinttermB(\br) &= \exp\left( -\frac{\cohwidth^2 k^2 \norm{ \br }_2^2}{2 d^2} \right),
\label{eqn:PropagationIntegralCohB}
\end{align}
$\lambda$ is the wavelength, $k=2\pi / \lambda$ is the wave number, and the Fourier transform is 2D with regard to the $x$ and $y$ components of $\br'$
(see Appendix~\ref{Sec:CoherenceDerivations} for derivation).
The variable $\br'$ indicates spatial position in the object plane, whereas $\br$ indicates spatial position along the wall.

Because $\brho$ appears in the argument to the Fourier transform of \eqref{eqn:PropagationIntegralCoh}, a natural way to measure the coherence function $\Coh_d(\br,\brho)$ is along the $\rho_x$ and $\rho_y$ axes with $\br$ fixed, i.e. measure a 2D slice of the 4D coherence function.
We will refer to a set of measurements along this slice as a coherence sample.
An example plot of such a coherence sample is shown in Fig.~\ref{Fig:FreeSpacePropMeas}(b), with a detail zoom shown in (c). Here, $\br=(-0.4~\text{m},-0.4~\text{m})$ is fixed, and the plot is over $\brho$.
The simulation parameters are $\lambda=525~\mu$m and $\cohwidth=2.5~\mu$m.

For the interaction with the wall, the angular spread of photons can be assumed to be governed by a Gaussian function \cite{Liu:15,Shen:17}. The standard deviation of the angular spread along the $x$ and $y$ axes is $\wallsdvec = (\wallsd_x,\wallsd_y)$.
The geometry of the scene is such that the angles of incidence and reflection are fairly close, which results in a specular reflection due to surface scattering.
Due to the paraxial nature of the incident waves, coupled with the narrow spread of the specular reflection \cite{TAVASSOLY2004252}, we can use the approximation
\begin{align} \label{eqn:scatteredCoh}
\Coh_S(\br, \brho)
&\approx
\propinttermscatter(\brho) \Coh_d(\br, \brho)
\end{align}
where 
\begin{align} \label{eqn:scatteredCohA}
\propinttermscatter(\brho) &=  \wallsd_x^2 \wallsd_y^2 \exp\left( -\frac{\wallsd_x^2 \rho_x^2}{2} -\frac{\wallsd_y^2 \rho_y^2}{2} \right),
\end{align}
and the intensity of the scattered field
\begin{align} \label{eqn:PropagationIntegralInt}
\Intensity_S(\br)
&= \frac{2 \pi \cohwidth^2}{\lambda^2 d^2}
\, \propinttermB(\br)
\star
\sourceimg(\br),
\end{align}
where $\star$ is the 2D convolution operator.
Fig.~\ref{Fig:FreeSpacePropMeas}(d) shows the results of wall scattering with parameters $\wallsdvec=(1~\micron,6~\micron)$.

To achieve spatial diversity, a full reconstruction will typically require a collection of 2D coherence samples, each centered at a different $\br$.  An example collection of 49 samples is given in Fig.~\ref{Fig:FreeSpacePropMeas}(e) showing the coherence incident to the wall, with the $\br$ falling onto a $7 \times 7$ grid. The corresponding scattered coherence functions are shown in Fig.~\ref{Fig:FreeSpacePropMeas}(f).

We remark that while the 2D intensity function constitutes a slice of the 4D coherence function, cameras used to measure intensity differ from devices used to measure coherence, therefore they are commonly considered as different modalities.

\section{Non-line-of-sight Image reconstruction} \label{section:ProposedMethod}

In this section, we turn to the problem of reconstructing the opacity profile of the object. 
This 2D profile is represented in discretized form by matrix $\sourceimgmat$, which has vectorized form $\sourceimgvec = \vectbrack{ \sourceimgmat }$.
Matrix $\sourceimgmat$ is formed by sampling the opacity profile on a uniform grid over the finite support of the profile.
First, we consider reconstruction using intensity-only measurements in the presence of ambient light from secondary sources. Then, leveraging the physical model for spatial coherence introduced in Section \ref{section:PhysicalModel}, we develop the reconstruction framework using coherence measurements. Finally, we define the complete problem in which we fuse information from both modalities and exploit the natural sparsity of the object's profile.

\subsection{Intensity Measurements}

The intensity pattern on the wall may be measured using a variety of readily available devices. For example, if intensity variations are strong enough, a simple Charge-Coupled Device (CCD) camera with a suitable lens may be used.
At the other extreme, a device such as an Electron Multiplying CCD (EMCCD) can distinguish minute intensity variations, due to the camera's high single photon sensitivity.

We define the intensity measurement matrix $\sensematI$, which samples the scattered intensity function $\Intensity_S(\br)$ in \eqref{eqn:PropagationIntegralInt} at the wall. Hence, in discretized form
\begin{align} \label{eqn:discretemeasmatI}
\sensematI \sourceimgvec = \frac{2 \pi \cohwidth^2}{\lambda^2 d^2}
\,
\vectbrack{
\propinttermBmat
\star
\left( \bone - \sourceimgmat \right)
}
\end{align}
where $\propinttermBmat$ is the discretized Gaussian kernel $\propinttermB(\br')$ defined in \eqref{eqn:PropagationIntegralCohB}.
Because $\sourceimgmat$ is an opacity profile, the intensity in the object plane takes the form $\bone - \sourceimgmat$, where the $\bone$ term represents the light incident on the object immediately prior to obstruction.

In the experiments, we implement \eqref{eqn:discretemeasmatI} using a linear convolution, i.e., elements outside the boundaries of the domain of $\br'$ are set to zero. This operation is performed through the use of convolution matrices such that the grids of $\br$ and $\br'$ may be different. If the grids are the same, we could also use the Fast Fourier Transform (FFT) to perform a fast circular convolution.

Therefore, to recover an estimation of the object profile $\sourceimgvec$ from intensity measurements (see Section~\ref{sec:NullSpace} for a discussion of the null space), we formulate the convex program  
\begin{align} \label{eqn:optproblemInt}
\min_{\sourceimgvec,\ambientcoeff} \norm{\sensematI \sourceimgvec  + \ambientcoeff \ambientvec - \measurematI}_2^2,
\end{align}
where $\measurematI$ is the measurement vector.
This formulation includes a free coefficient $\ambientcoeff$ along with an associated vector $\ambientvec$ modeling the ambient light. Specifically, vector $\ambientvec$ captures the spatial intensity distribution of the ambient light on the wall and
the coefficient $\ambientcoeff$ represents its magnitude.
Here, we set $\ambientvec=\bone$, i.e., the ambient light blankets the wall with constant intensity.

While this problem may be successful if a clear shadow is discernible, two major factors limit its effectiveness.
First, the shadow will be faint if there is significant ambient light present.  Although the shadow can be measured with sensitive cameras, the Signal-to-Noise Ratio (SNR) falls as the amount of ambient light increases.
Second, if the coherence of the light sources is low, the edges of the shadow will be indistinct due to diffraction, making the reconstruction ill-posed; this effect can be seen as a manifestation of the convolution in \eqref{eqn:PropagationIntegralInt}.

\subsection{Coherence Measurements}

To address the aforementioned limitations of the intensity-based approach, we develop a framework for reconstruction from coherence measurements next.

As described in the introduction, an increasing number of techniques have been developed for capturing coherence information.
An example of practical measurements matching the requirements of our approach can be found in \cite{Batarseh2018}, which makes use of a Dual Phase Sagnac Interferometer (DuPSaI).

We define the coherence measurement matrix $\sensematC{\br}$, which samples the scattered coherence function along the $\rho_x$ and $\rho_y$ axes at a fixed $\br$. Obtaining a discretized form of the function in \eqref{eqn:scatteredCoh}, we can write
\begin{align} \label{eqn:discretemeasmatC}
\sensematC{\br} \sourceimgvec = 
\vectbrack{
\propinttermscattermat \hadprod \propinttermAmat_\br \hadprod \left( \bF_1 \left[ \propinttermBmat_\br \hadprod \left( \bone - \sourceimgmat \right) \right] \bF_2 \right) }.
\end{align}
Matrix $\propinttermscattermat$ is the discretized form of the function $\propinttermscatter(\brho)$ defined in \eqref{eqn:scatteredCohA}, which represents the scattering effects of the wall.
Matrix $\propinttermAmat_\br$ is the discretized form of the function $\propinttermA (\br,\brho)$ defined in \eqref{eqn:PropagationIntegralCohA}, which is one component of the free-space propagation operator.
Both $\propinttermscattermat$ and $\propinttermAmat_\br$ are discretized along the $\rho_x$ and $\rho_y$ axes using the same set of points as $\sensematC{\br}$, with $\propinttermAmat_\br$ using the same fixed $\br$ position as $\sensematC{\br}$.
The other component of the free-space propagation operator is matrix $\propinttermBmat_\br$, which discretizes the function $\propinttermB$ defined in \eqref{eqn:PropagationIntegralCohB}.
Specifically, this matrix contains samples of $\propinttermB(\br-\br')$, with $\br$ fixed, and $\br'$ falling on the same discrete grid as $\sourceimgmat$.

Calculation using the measurement matrix \eqref{eqn:discretemeasmatC}
admits a tractable form, requiring only element-wise products and Fourier transforms, which may be implemented using the FFT.

The measurement vector corresponding to the coherence sample at $\sensematC{\br}$ is labeled $\measurematC{\br}$. We define the set $\measRsetC$ containing the values of $\br$ at which the full collection of coherence samples are made. To perform the reconstruction using coherence measurements, we consider the least squares formulation,
\begin{align} \label{eqn:optproblemC}
\min_{\sourceimgvec} \sum_{\br \in \measRsetC} \norm{\sensematC{\br} \sourceimgvec - \measurematC{\br}}_2^2.
\end{align}

A major factor influencing the quality of the coherence measurements are the geometry and characteristics of the wall which determine the amount of scattering.  Because these factors may vary depending on spatial position along the wall, the different sets of measurements $\measurematC{\br}$ within the collection may vary in their quality, or some may be unusable.  We will explore such a scenario in Section~\ref{sec:FusionResults}.

Given the geometry of the scene, the ambient light that reaches the detector will necessarily result from diffuse scattering (i.e., specularly reflected ambient light from secondary sources will not reach the detector due to unequal angles of incidence and reflection).
Because there is a Fourier transform relationship between scattered photon angle and coherence (see Appendix~\ref{Sec:CoherenceDerivations} for more details), the large angle diffuse spread in the ambient light introduces a narrow peak in the coherence function at $\brho=\bzero$ \cite{Shen:17,Shen_FiO:17}.
Recalling the relationship between intensity and coherence $\Intensity(\br) = \Coh(\br, \bzero)$, we can see that the peak exactly coincides with the intensity measurements.  Therefore, the ambient light tends to dominate the intensity measurements and obscure the shadow.
On the other hand, this diffusely scattered ambient light has little effect on the coherence function away from $\brho=\bzero$, where the specular component of reflection (containing information about the object) dominates.
For this reason, spatial coherence coordinates for which
$\norm{\brho}_2 < \cohzerorad$ are excluded.
We remark that unlike \eqref{eqn:optproblemInt}, this exclusion obviates the need for an ambient term for the coherence measurements in the formulation of \eqref{eqn:optproblemC}.

\subsection{Fusion Framework}

As mentioned in the previous sections, it is possible that one or another modality may be of a lower quality, and therefore it is advantageous to use both intensity and coherence modalities in the same reconstruction.

Additionally, the profile $\sourceimgvec$ is likely to admit a sparse representation $\bx$ in a particular basis $\sparsemat$.
Here, we use the two-dimensional Discrete Cosine Transform (DCT) as the sparsifying basis $\sparsemat$ (in which it is well established that natural images possess a sparse representation \cite{Taubman:02}), however, another basis such as a wavelet basis could also be used.
As such, the object profile can be expressed as $\sourceimgvec = \sparsemat \bx$.
We then include $\norm{\bx}_1$ as a regularization term to promote sparsity in the reconstruction, where the $\ell_1$-norm is a convex relaxation of the $\ell_0$-norm \cite{BoydVandenberghe:04}.

To fuse information from both modalities and exploit the sparsity of the opacity profile in $\sparsemat$, we can readily formulate the convex program
\begin{align} \label{eqn:optproblemfull}
\min_{\bx,\ambientcoeff}
\,
&\sparseweight \norm{\bx}_1
+ \norm{\sensematI \sparsemat \bx  + \ambientcoeff \ambientvec - \measurematI}_2^2
\nonumber \\
&+ \measureweight \sum_{\br \in \measRsetC} \norm{\sensematC{\br} \sparsemat \bx - \measurematC{\br}}_2^2,
\end{align}
where $\sparseweight$ and $\measureweight$ are used to balance the objectives.

\subsection{Algorithm}
To solve \eqref{eqn:optproblemfull}, we propose an iterative algorithm based on the ADMM approach first introduced in \cite{Boyd:2011:DOS:2185815.2185816}. This algorithm performs a dual ascent using the Augmented Lagrangian \cite{Hestenes1969}, which can be written as
\begin{align}
&L_\penaltyweight( \bx,\ambientcoeff, \bz,\lagrangemult )
= \sparseweight \norm{ \bz }_1
+ \norm{\sensematI \sparsemat \bx  + \ambientcoeff \ambientvec - \measurematI}_2^2
\nonumber \\
&+ \measureweight \sum_{\br \in \measRsetC} \norm{\sensematC{\br} \sparsemat \bx - \measurematC{\br}}_2^2
+ \realpart{\lagrangemult^* (\bx-\bz)} + \frac{\penaltyweight}{2} \norm{ \bx-\bz }_2^2
\nonumber
\end{align}
where $\lagrangemult$ is the Lagrange multiplier.
We solve the minimization using the following updates at each step $k$:
\begin{align}
\bx^{k+1}, \ambientcoeff^{k+1} &= \argmin_{\bx,\ambientcoeff} L_\penaltyweight \left( \bx,\ambientcoeff, \bz^k, \lagrangemult^k \right)
\label{eqn:admm_xmin},
\\
\bz^{k+1} &= \argmin_\bz L_\penaltyweight \left( \bx^{k+1},\ambientcoeff^{k+1}, \bz, \lagrangemult^k \right)
\label{eqn:admm_zmin},
\\
\lagrangemult^{k+1} &= \lagrangemult^k - \penaltyweight (\bz-\bx),
\end{align}
where the initial values $\bx^0,\ambientcoeff^0,\bz^0,\lagrangemult^0$ are zero.
The stopping criteria consist of thresholds placed on the residuals \cite{Boyd:2011:DOS:2185815.2185816}.  Specifically, the algorithm stops if the norm of the primal residual $\norm{\bx^k-\bz^k}_2 < \stopthreshprime$ and the norm of the dual residual $\norm{\penaltyweight(\bx^{k+1}-\bx^{k})}_2 < \stopthreshdual$.
Here, $\stopthreshprime=0.5$ and $\stopthreshdual = 10^{-6}$.

Details regarding the calculation of the $\bx$ and $\bz$ update steps are given in Appendix~\ref{sec:optalgdetails}.

\subsection{Mapping of Null Space} \label{sec:NullSpace}

Due to various factors in the propagation and scattering process, the measurement matrices $\sensematI$ and $\sensematC{\br}$ will typically possess a null space.  We use the general notation $\sensemat_i$ to refer to the $i\textsuperscript{th}$ measurement matrix, which may take the form of $\sensematI$ or $\sensematC{\br}$, depending on the enumeration order of the matrices.

We can characterize the null space associated with measurement $i$ as follows.
The degree of coherence between the $j\textsuperscript{th}$ element of the object profile $\sourceimgvec_j$ and the measurement can be quantified by
$\sensemapspatial_i(j) = \norm{ \sensemat_i \be_{j} }_2$.
If $\sensemapspatial_i(j)$ is close to zero, i.e. the SNR is very small, the element is considered to be in the null space of the measurement.

Similarly, we can look at the degree of coherence in the sparse domain using a similar operator
$\sensemapfreq_i(j) = \norm{ \sensemat_i \sparsemat \be_{j} }_2$.
The null space map may be especially useful when an explicit model is not known, for example in data-driven approaches.

\subsection{Sample Weighting} \label{sec:SampleLevelWeight}
It may improve the results if we can exclude certain measurements from the reconstruction rather than give equal weight to all measurements in the samples.
To this end, we can substitute a weighted norm $\norm{\cdot}_{\normweight}$ in place of any of the Euclidean norms $\norm{\cdot}_2$ in \eqref{eqn:optproblemfull}.

If the noise is known, the sample weight vector for the $i\textsuperscript{th}$ measurement can be constructed using the decision metric
\begin{align} \label{eqn:sampleweightdecider}
\normweight_i(j) = \left(
\frac{1}{\measnoise_i}
\quad
\underset{0}{\overset{1}{\gtrless}}
\quad
\frac{\fineweightcal}{N-1}
\sum_{\stackrel{k=1}{k \ne i}}^N \frac{\norm{\sensemat_k \sensemat_i^* \be_j}_2}{\measnoise_k}
\right),
\end{align}
where $j$ is the sample number, $\measnoise_i$ is the noise level present in measurement $i$, and $\fineweightcal$ is a calibration constant.
This is a metric similar to the Transform Point Spread Function found in \cite{doi:10.1002/mrm.21391}.  For a given measurement sample, this metric finds other samples which are coherent with the same image pixels.
A given sample will be included in the optimization if it has a higher SNR than the other measurements.

\subsection{Extensions}

Here, we comment on possible extensions to the framework.

We are not constrained to problems in which the object is blocking light, but can also work in reflective scenarios.  This can be accomplished by redefining $\sourceimgmat$ as the reflectivity rather than opacity of the object, and making the simple substitution $\bone - \sourceimgmat \rightarrow \sourceimgmat$ in \eqref{eqn:discretemeasmatI} and \eqref{eqn:discretemeasmatC}.

The problem \eqref{eqn:optproblemfull} includes a single weight $\measureweight$ associated with the measurements.  We may instead associate a weight coefficient with each measurement matrix in \eqref{eqn:optproblemfull}.
These could be adjusted along a continuum to control the impact of particular samples.
If the magnitudes of measurements are significantly different, these weights can maintain balance, e.g., by setting $\measureweight_i = 1/\norm{\measuremat_i}_2^2$.
If there is Gaussian noise in the measurements with known magnitude, the Bayesian Compressive Sensing methodology can be used \cite{4524050}.

Another possible extension to the optimization problem is to incorporate an auto-scaling coefficient, e.g., to handle cases when the magnitude of measurements from different modalities are not calibrated to the same scale.
To this end, we can add a scaling coefficient $\measscale$ to some of the measurements by making the substitution
$\measuremat_i \rightarrow \measscale \measuremat_i$, and updating $\measscale$ in step \eqref{eqn:admm_xmin}.
With this modification, the problem \eqref{eqn:optproblemfull} remains convex.

\section{Numerical Results} 
\label{section:Results}

\begin{figure}[ht]
\centering
\includegraphics[scale=1]{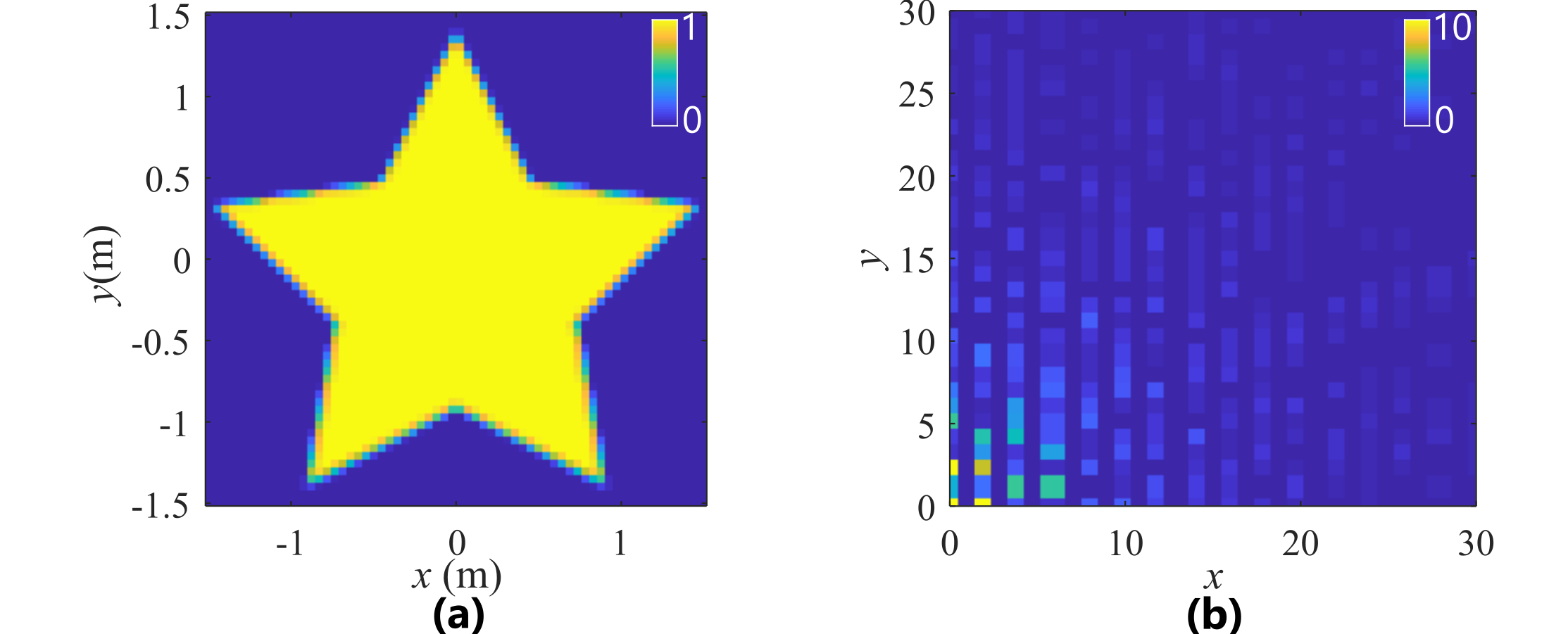}
\vspace{-0.7cm}
\caption{
(a) Actual opacity used at object plane in forward model for all results.
(b) Corresponding DCT.  The color range for the DCT plot is restricted to $[0,10]$ to highlight components with smaller magnitudes.
}
\label{Fig:SourceImage}
\end{figure}

We now present examples demonstrating the proposed method laid out in Section~\ref{section:ProposedMethod} and making use of optimization problem \eqref{eqn:optproblemfull}.
In all examples, the opacity profile of the actual object is as shown in Fig.~\ref{Fig:SourceImage}(a) with corresponding DCT in (b).

For simulated measurements, the source intensity function $I(\br)$ used in the forward model is as shown in the diagram of Fig.~\ref{Fig:FreeSpacePropMeas}(a) (left side), with the function extended by ones to $x,y \in (\pm 6~\text{m})$, thus representing an opaque star object surrounded by a plane of light.  The extension of the function is required to properly model the significant spreading of the light after being emitted from the physical light sources and before being obstructed by the object.

Additive-white-Gaussian-noise (AWGN) with standard deviation (SD) $\measnoiseI$ is added to the intensity samples, and complex AWGN with SD $\measnoiseC$ is added to the coherence samples.

The following parameters are used in all results: $\lambda=525~\mu$m, $d=6$~m, $\cohzerorad = 1\,\mu$m, $\penaltyweight=5 \times 10^{-3}$, and $\measureweight=1$. The intensity image of the wall has resolution $101 \times 101$ pixels with domain $r_x,r_y \in$ [$\pm 2$~m]. Unless otherwise specified, the coherence measurements have resolution $51 \times 51$ pixels (with the domain of $\brho$ varying depending on the example).
A constant value of 100 is added to all intensity measurements to model ambient light; this offset will be absorbed by the coefficient $\ambientcoeff$ in \eqref{eqn:optproblemfull}.

\subsection{Non-line-of-sight Imaging}

We first demonstrate the potential of spatial coherence measurements to enable passive NLOS imaging when no shadow information is available.  Two reconstructions are included, each with wall scattering parameters set at opposite extremes.

In this example, coherence measurements are made on the same spatial grid as shown in Fig.~\ref{Fig:FreeSpacePropMeas}(f).
The simulation parameters are $\cohwidth=5~\mu$m, $\measnoiseC = 10^{-3}$, $\sparseweight=0$, and the coherence measurements are over domain $\rho_x,\rho_y \in$ [$\pm 15~\mu$m].

\begin{figure}[ht]
\centering
\includegraphics[scale=1]{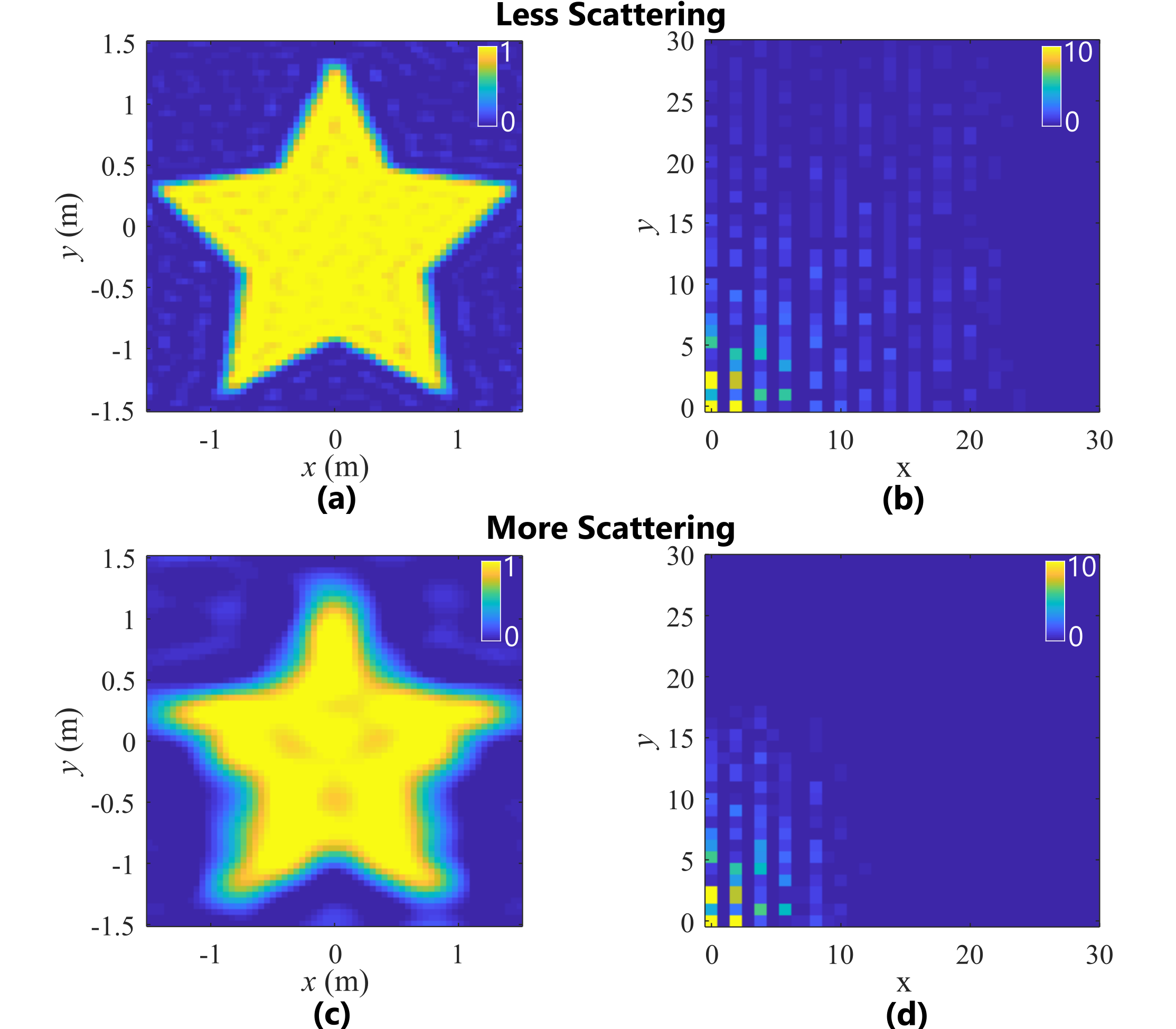}
\vspace{-0.7cm}
\caption{
Results of NLOS object reconstruction using coherence only.
The top half shows the reconstructed (a) image and (b) DCT for a wall that has relatively little scattering with $\wallsdvec=(3~\micron,18~\micron)$.  
(c) and (d) show the corresponding plots for a wall with more scattering where $\wallsdvec=(0.25~\micron,1.5~\micron)$.
Pixels in the reconstructed images with value $>1$ are set to one and values $<0$ are set to zero
}
\label{Fig:Results_CohOnly}
\end{figure}

Fig.~\ref{Fig:Results_CohOnly}(a) and (b) shows the reconstructed image and DCT for a wall with relatively little scattering, where the scattering parameters are set to $\wallsdvec=(3~\micron,18~\micron)$.

For comparison purposes, pixels in the reconstructed images with value $>1$ are set to one and values $<0$ are set to zero, a practice which will be used for the remainder of this section.

Fig.~\ref{Fig:Results_CohOnly}(c) and (d) show the results for a wall that introduces more scattering with parameters $\wallsdvec=(0.25~\micron,1.5~\micron)$.
The DCTs clearly show that the scattering of the wall acts as a low-pass filter, with increased scattering leading to more filtering.

\subsection{Fusion of Intensity and Coherence Measurements} \label{sec:FusionResults}

As demonstrated in the next example, by fusing intensity and coherence measurements, a better reconstruction can be made as compared to using either modality alone.

The simulation parameters used in this example are $\cohwidth=2.5~\mu$m, $\measnoiseI = 5 \times 10^{-2}$, $\measnoiseC = 10^{-2}$, $\wallsdvec=(2~\micron,6~\micron)$, $\sparseweight=10^{-3}$, and $\rho_x,\rho_y \in$ [$\pm 10~\mu$m].

\begin{figure}[ht]
\centering
\includegraphics[scale=1]{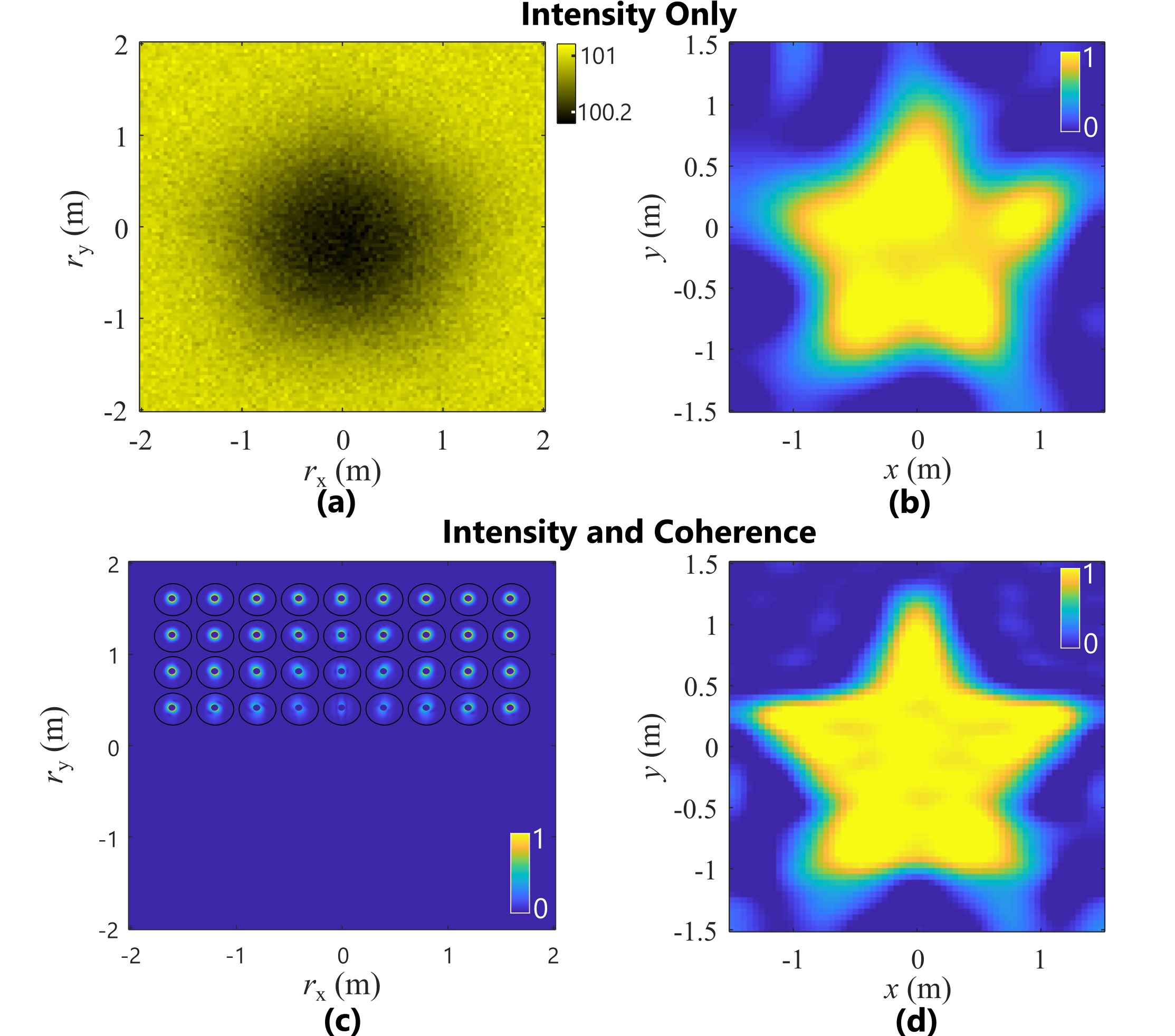}
\vspace{-0.7cm}
\caption{
Fusion results.
(a) Shows an intensity sample, and (b) shows the reconstruction using this sample alone.
(c) Shows an additional measurement of scattered coherence, each sample having plot radius $5.2~\mu$m.  The measurements only cover part of the wall.
(d) Shows the reconstruction when both the intensity and coherence measurements are used.
}
\label{Fig:MainFusionExample}
\end{figure}

First, Fig.~\ref{Fig:MainFusionExample}(a) shows an intensity sample.  Note that the color range of the intensity plot has been constrained to a narrow range to clearly show the shadow.  The light is not coherent enough to reveal the edges of the star.
Fig.~\ref{Fig:MainFusionExample}(b) shows the reconstructions results when only this sample is used.

Next, the coherence samples shown in Fig.~\ref{Fig:MainFusionExample}(c) are also included in the reconstruction to augment the intensity measurements.  Fig.~\ref{Fig:MainFusionExample}(d) shows the improved results.  In the top half of the reconstruction, the coherence measurements contain more information about the high frequency components of the object profile and therefore dominate the reconstruction providing sharper edges.  However, because these coherence measurements only cover the top half of the wall, the intensity contains more information about the bottom half of the object, albeit only at lower frequencies thus resulting in less definition.

We will now provide some insight into the improvements which have been made based on Section~\ref{sec:NullSpace}.

\begin{figure}[t!]
\centering
\includegraphics[scale=1]{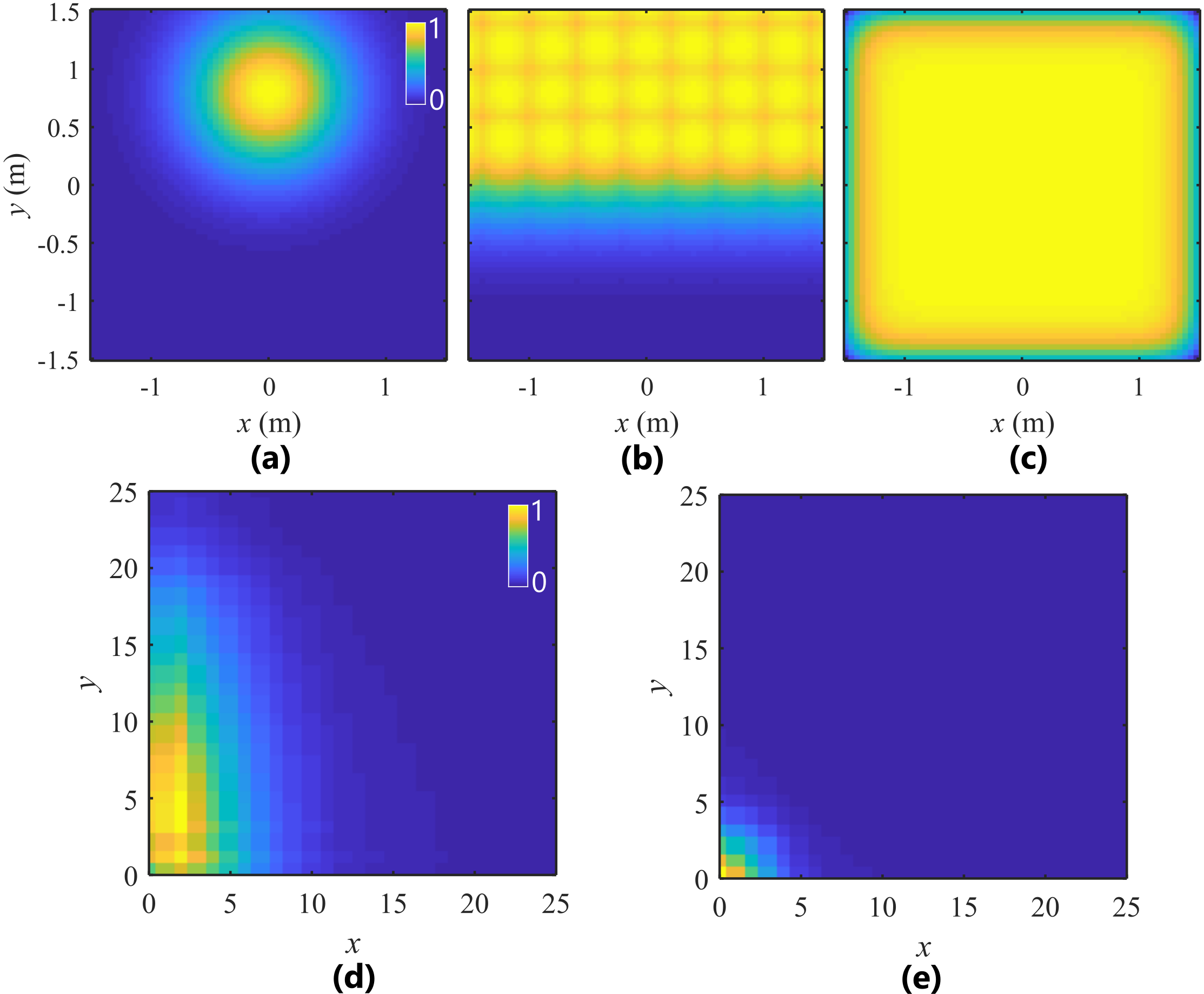}
\vspace{-0.7cm}
\caption{
Null spaces for measurements (small values indicate an element is in the null space).  Null spaces in image basis of (a) single coherence sample, (b) all coherence samples, and (c) intensity sample.  Null space in DCT basis of (d) all coherence samples and (e) intensity sample.
All color scales are normalized to their respective maximum values.
}
\label{Fig:FusionCoherence}
\end{figure}

First, the spatial limitation inherent in coherence measurements is demonstrated.
This limitation comes from the multiplication by the Gaussian term $\propinttermB(\br)$ in \eqref{eqn:PropagationIntegralCoh}. In the following discussion, we denote the index of the intensity sample as $\measidxsetI$, and the index set of coherence samples as $\measidxsetC$.
In Fig.~\ref{Fig:FusionCoherence}(a), we show $\sensemapspatialvec_{i}$ for a single coherence sample located at $\br=(0,0.8~\text{m})$.
Fig.~\ref{Fig:FusionCoherence}(b) shows $\max_{i \in \measidxsetC} \sensemapspatialvec_{i}$, which returns a vector containing the most coherent coherence measurements with each pixel.
This is the combined effect of all coherence samples, clearly demonstrating that more samples allow more spatial coverage.  In contrast, Fig.~\ref{Fig:FusionCoherence}(c) shows $\max_{i \in (\measidxsetI \cup \measidxsetC) } \sensemapspatialvec_{i}$, demonstrating that when all  coherence measurements are used together with intensity measurements, virtually the entire object profile is covered.

We can perform a similar analysis in the sparse DCT domain.
Fig.~\ref{Fig:FusionCoherence}(d) shows $\max_{i \in \measidxsetC} \sensemapfreqvec_{i}$, which is the combined effect of the coherence samples in the sparse basis, and  Fig.~\ref{Fig:FusionCoherence}(e) shows $\sensemapfreqvec_{\measidxsetI}$, which is the effect of the intensity measurements in the sparse basis.
It can be seen that the coherence measurements have a stronger correlation with the high frequency components, explaining why the top half of Fig.~\ref{Fig:MainFusionExample}(b) has improved edges over the bottom half.
The low pass filtering in the intensity measurements comes from the convolution in \eqref{eqn:PropagationIntegralInt} due to diffraction, whereas the filtering in the coherence measurements comes from wall scattering.

\subsection{Improved Fusion using Sample Weighting}

In some cases, simply adding new measurements is insufficient.  Because of noise levels, while certain parts of the reconstruction will improve, other parts will degrade.  In this cases being able to exclude individual measurements as described in Section~\ref{sec:SampleLevelWeight} may resolve the issue.  We now provide such an example.

\begin{figure}[ht]
\centering
\includegraphics[scale=1]{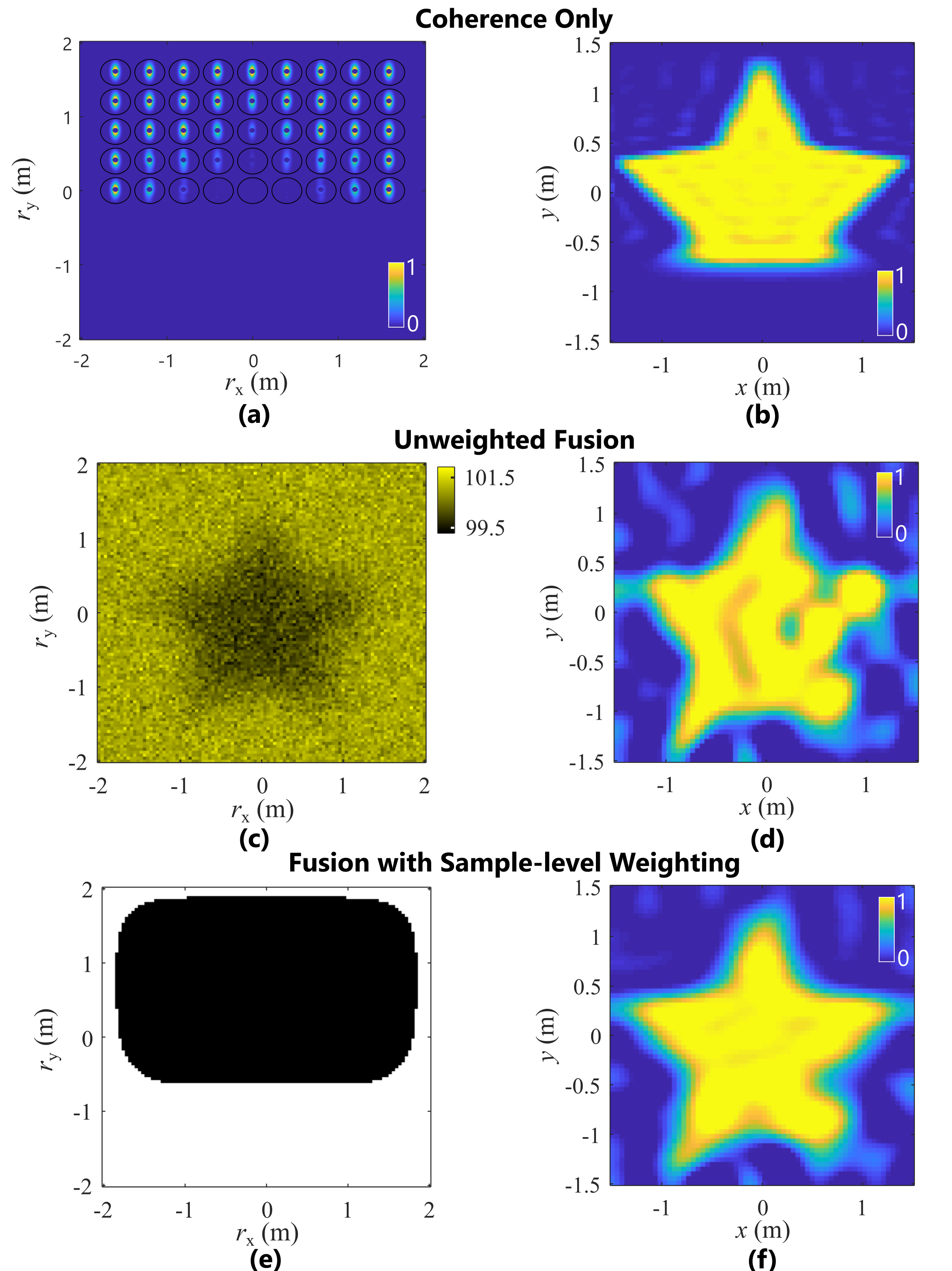}
\vspace{-0.7cm}
\caption{
Example of sample weighting.
(a) shows a set of coherence samples, each with plot radius $5.4~\mu$m, and
(b) the corresponding reconstruction.  As in the example of Fig. \ref{Fig:MainFusionExample}, the coherence measurements only provide a reconstruction of the top half of the image.
(c) An intensity sample, which is fused with the coherence samples to create reconstruction (d).  Due to noise in the intensity measurements, the quality of the reconstruction is poorer.
(e) The sample weights for the intensity measurement as calculated using \eqref{eqn:sampleweightdecider}.  White regions indicate intensity samples which are included in the reconstruction, i.e. measurements $j$ for which $\normweightvec_{\measidxsetI}(j)=1$, and black regions indicate exclusions, i.e. for which $\normweightvec_{\measidxsetI}(j)=0$.
(f) Reconstruction from the same intensity and coherence samples, but using the weights shown in (e).
}
\label{Fig:SampleWeights}
\end{figure}

The simulation parameters used in this example are $\cohwidth=5~\mu$m, $\measnoiseI = 0.25$, $\measnoiseC = 10^{-4}$, $\wallsdvec=(1~\micron,6~\micron)$, $\brho_x,\brho_y \in$ [$\pm 15~\mu$m], and $\fineweightcal=0.25$.  For Fig.~\ref{Fig:SampleWeights}(b), $\sparseweight=0$ and for Fig.~\ref{Fig:SampleWeights}(d) and (f), $\sparseweight=1.5 \times 10^{-2}$.

Coherence measurements and the associated reconstruction are shown in Fig. \ref{Fig:SampleWeights}(a) and (b) respectively.
In these panels we do not use regularization, since the measurements lack noise.

If the intensity sample shown in Fig.~\ref{Fig:SampleWeights}(c) is also used in addition to the coherence measurements, the results in Fig.~\ref{Fig:SampleWeights}(d) are obtained.  Here, sparsity regularization is used due to noise in the intensity measurements.  Although the bottom half of the object is now visible in the reconstruction, the top half has degraded due to the intensity noise.

To resolve this problem, we calculate sample weights for the intensity measurement using \eqref{eqn:sampleweightdecider}.  The results are shown in Fig.~\ref{Fig:SampleWeights}(e) with black representing zeros (excluded intensity samples) and white representing ones (included samples).

The result of the reconstruction using these weights is shown in Fig.~\ref{Fig:SampleWeights}(f), where the top half can be seen to improve.  Note that because we are regularizing in the frequency domain, noise which is \textit{spatially} isolated to a particular section of the image will be coupled to other noise-free regions, and thus the top half is not ideal as possible.  Using a wavelet basis may eliminate this issue.

\subsection{Sparsity}

Fig.~\ref{Fig:SourceImage}(b) confirms that the DCT of this object profile is approximately sparse (disregarding the small high frequency components).
At the same time, noise in the measurements tends to introduce relatively large high frequency components into the reconstruction.  Therefore, one use for the sparsity regularizer in \eqref{eqn:optproblemfull} is to serve as a de-noising tool.

\begin{figure}[ht]
\centering
\includegraphics[scale=1]{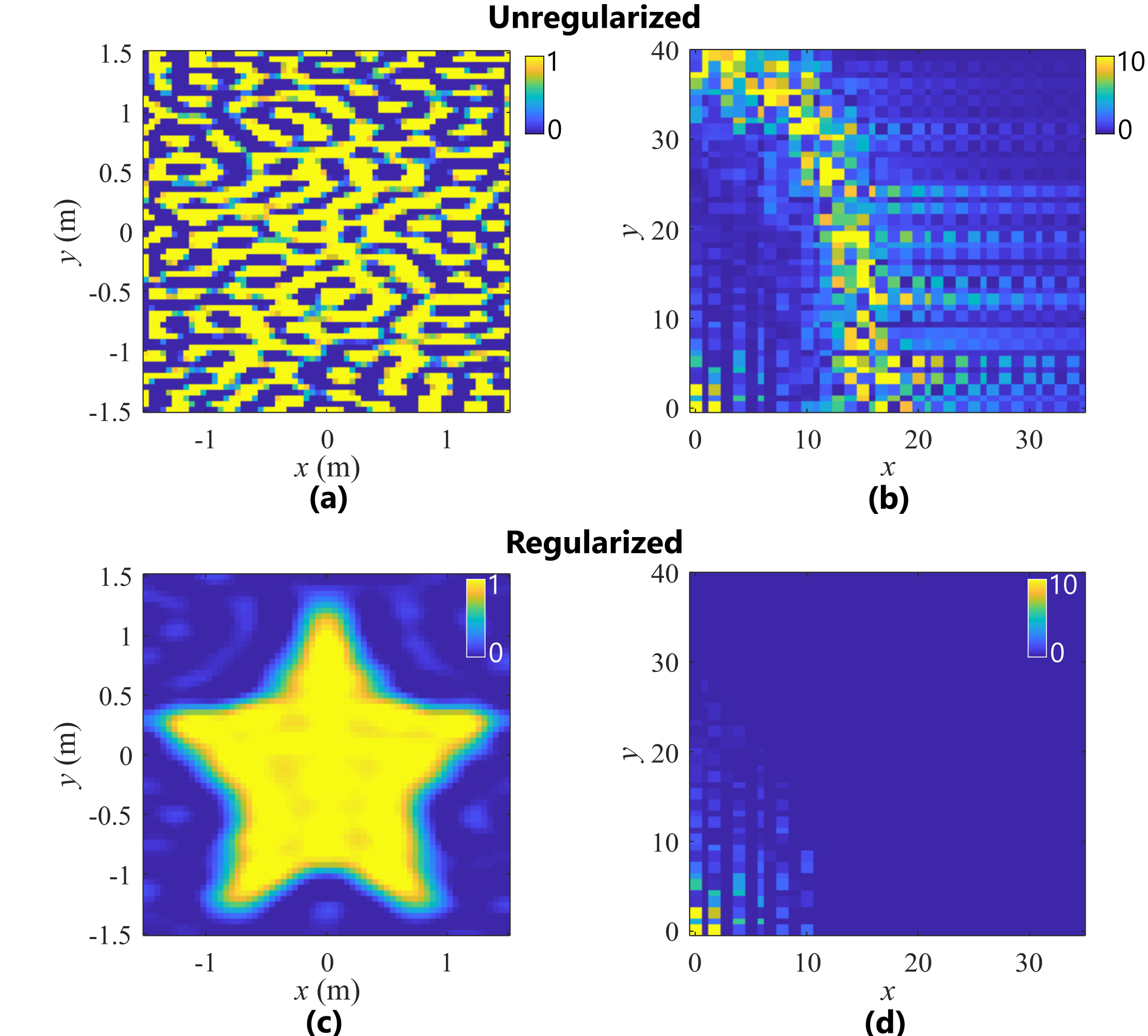}
\vspace{-0.7cm}
\caption{
Demonstration of $\ell_1$ norm regularization for noise reduction.  In this example, only coherence measurements are used.
(a) and (c) shows the reconstructed object profiles.
(b) and (d) shows the corresponding DCTs.
}
\label{Fig:Sparsity}
\end{figure}

In Fig.~\ref{Fig:Sparsity}(a) we show the result of a reconstruction using noisy coherence measurements where no regularization is used, i.e. $\sparseweight=0$.
As shown in Fig.~\ref{Fig:Sparsity}(b), the noise appears mostly in the high frequency components of the DCT. 
Fig.~\ref{Fig:Sparsity}(c) and (d) show the improved results when sparsity is enforced using $\sparseweight=5 \times 10^{-4}$.

The coherence measurements are at the same spatial locations as shown in Fig.~\ref{Fig:FreeSpacePropMeas}(f).
The simulation parameters are $\cohwidth=5~\mu$m, $\measnoiseC = 10^{-2}$, $\wallsdvec=(1~\micron,6~\micron)$ and $\rho_x,\rho_y \in$ [$\pm 15~\mu$m] with a resolution of $25 \times 25$.

In Table~\ref{tab:tableerror_coherence}, we repeat this experiment using the same parameters, except varying the noise levels and $\sparseweight$ values. Ten trials are performed at each setting, and the average and SD of the resulting Mean Square Error (MSE) are shown.

Likewise, Table~\ref{tab:tableerror_intensity} shows the results using only intensity measurements (and no coherence measurements). Here, the coherence level used for the forward model is $\cohwidth=2.5~\mu$m (to reduce the distinctness of the shadow).

For each noise level (column), the minimum error is bolded. We can see in both tables that a larger noise level requires a larger value of $\sparseweight$ to achieve minimal MSE. The errors in the bottom row are roughly equal for all noise levels: beyond a certain threshold of $\sparseweight$, the estimates only contain low frequency components and are nearly identical.

\begin{table}[t!]
  \centering
  \caption{\bf MSE of coherence-only measurements (average and standard deviation) }
  \begin{tabular}{c cccc}
  \hline
  \multirow{2}{*}{$\sparseweight$}	& \multicolumn{4}{c}{Noise ($\measnoiseC$)} \\
  \cmidrule(lr){2-5}
 & $0$ & $0.01$ & $0.05$ & $0.1$ \\
  \hline
$0$ & $\mathbf{0.008}$ & $5.334 \pm 1.111$ & $126 \pm 19$ & $621 \pm 76$ \\
$0.0005$ & $0.014$ & $\mathbf{0.015 \pm 0.001}$ & $0.59 \pm 0.08$ & $9.35 \pm 2.23$ \\
$0.005$ & $0.020$ & $0.020 \pm 0.000$ & $\mathbf{0.03 \pm 0.00}$ & $0.12 \pm 0.02$ \\
$0.05$ & $0.041$ & $0.041 \pm 0.000$ & $0.04 \pm 0.00$ & $\mathbf{0.04 \pm 0.00}$ \\
$0.5$ & $0.140$ & $0.140 \pm 0.000$ & $0.14 \pm 0.00$ & $0.14 \pm 0.00$ \\
    \hline
  \end{tabular}
  \label{tab:tableerror_coherence}
\end{table}

\begin{table}[t!]
  \centering
  \caption{\bf MSE of intensity-only measurements (average and standard deviation) }
  \begin{tabular}{c cccc}
  \hline
  \multirow{2}{*}{$\sparseweight$}	& \multicolumn{4}{c}{Noise ($\measnoiseI$)} \\
  \cmidrule(lr){2-5}
 & $0$ & $0.1$ & $0.5$ & $1$ \\
  \hline
$0$ & $\mathbf{0.04}$ & $0.09 \pm 0.02$ & $0.742 \pm 0.223$ & $3.019 \pm 1.817$ \\
$0.05$ & $0.06$ & $\mathbf{0.07 \pm 0.00}$ & $0.138 \pm 0.032$ & $1.146 \pm 0.532$ \\
$0.10$ & $0.08$ & $0.08 \pm 0.00$ & $0.089 \pm 0.014$ & $0.402 \pm 0.227$ \\
$0.50$ & $0.09$ & $0.09 \pm 0.00$ & $\mathbf{0.088 \pm 0.003}$ & $0.097 \pm 0.009$ \\
$1.00$ & $0.09$ & $0.09 \pm 0.00$ & $0.091 \pm 0.003$ & $\mathbf{0.096 \pm 0.006}$ \\
$5.00$ & $0.12$ & $0.12 \pm 0.00$ & $0.115 \pm 0.002$ & $0.123 \pm 0.009$ \\
    \hline
  \end{tabular}
  \label{tab:tableerror_intensity}
\end{table}

\section{Discussion} \label{section:Discussion}

Here, we considered the problem of passive NLOS imaging.  This theoretical study, based on reliable models, aimed at providing a framework for solving the inverse problem, fusing multi-modal measurements, and understanding the measurement operators.

The reconstruction algorithm can leverage intensity (i.e. shadow) information when available.
However, during the propagation process, intensity becomes blurred.  
As an alternative, we can use samples of spatial coherence, which retains information during the propagation process. However, scattering may significantly attenuate this surviving information.  By fusing the two modalities, all information available in both sets of measurements may be captured.  If some of the measurements have a large noise level, a decision algorithm was presented by which they may be excluded.

In our work, we assume the optical distance to be known.
In \cite{Batarseh2018}, a technique is provided for determining the optical distance using the phase of the measurements at different spatial positions along the wall, information which is readily available in the measurements we use here.
This estimation could be performed as a preprocessing step, prior to running our algorithm.
The estimation of depth in the presence of scatterers has also been studied previously \cite{8237525,Liu:15}, and those results may help here as well.

In our problem, we reconstruct a planar object profile.  An extension of this work would be to consider three-dimensional objects, for example as was done in \cite{7997902} and \cite{doi:10.1002/ima.1016}.

We conclude by noting that the measurement matrices and optimization problem, as well the sample weighting and null space characterization, are general in nature.
While only two modalities were presented here, other modalities could be easily incorporated as well.

\appendices

\section{Optimization Algorithm Details} \label{sec:optalgdetails}

We define $\sparsepropmat_i := \sensemat_i \sparsemat$, and let $\normweightmat_i = \diag \normweightvec_i$ be the weight matrix associated with the weighted norms (if sample weighting is not used, then $\normweightmat_i$ should be an identity matrix).

The minimization step \eqref{eqn:admm_zmin} takes the analytic form \cite{Boyd:2011:DOS:2185815.2185816}
\begin{align}
\bz^{k+1} &= \overline{S}_{\sparseweight/\penaltyweight}\left(\bz^k + \lagrangemult^k/\penaltyweight \right),
\end{align}
where the component-wise shrinkage operator is
\begin{align}
\overline{S}_a(x_i) = \max\left\{ 1-a/\abs{x_i}, 0 \right\} x_i.
\end{align}

For simplicity, in the following equations we use a single summation over all samples, rather than separating the intensity sample from the coherence samples as was done in \eqref{eqn:optproblemfull}.  Additionally, the weight coefficient has been indexed and moved inside the summation.  For coherence samples, i.e. where $i \in \measidxsetC$, the ambient vector is set to $\ambientvec_i=\bzero$.

We solve step \eqref{eqn:admm_xmin} using a gradient descent algorithm.
The gradients are
\begin{align}
\nabla_\bx L_\penaltyweight
&= 2 \sum_{i \in (\measidxsetI \cup \measidxsetC)} \measureweight_i \realpart{\sparsepropmat_i^* \normweightmat_i \sparsepropmat_i \bx - \sparsepropmat_i^* \normweightmat_i (\measurevec_i - \ambientcoeff \ambientvec_i)}
\nonumber \\
&\qquad\qquad + \penaltyweight \realpart{\bx - \left( \bz - \lagrangemult / \penaltyweight \right)} ,
\nonumber
\\
\nabla_{\ambientcoeff} L_\penaltyweight
&= 2 \sum_{i \in (\measidxsetI \cup \measidxsetC)} \measureweight_i \realpart{\ambientcoeff \ambientvec_i^* \normweightmat_i \ambientvec_i - \ambientvec_i^* \normweightmat_i (\measurevec_i - \sparsepropmat_i \bx)} .
\nonumber
\end{align}
The initial conditions for the gradient descent at step $k\!+\!1$ are the values calculated at the previous step, i.e., $\bx^k$ and $\ambientcoeff^k$.
The $j\textsuperscript{th}$ step of the gradient descent inner loop is chosen to minimize the quadratic interpolation at points $\bx^j - \quadpoint\left( \nabla_{\bx} L_\penaltyweight\right)$ and $\ambientcoeff^j-\quadpoint \left( \nabla_{\ambientcoeff} L_\penaltyweight\right) $, where $\quadpoint \in \{0.1,0.5,1\}$.
Let $f^j=L_\penaltyweight(\bx^{j},\ambientcoeff^{j}, \bz^k,\lagrangemult^k)$.
The descent algorithm stops when
$\norm{ f^{j+1} - f^{j} } / f^{j} < \gradstopthresh$.

For the $\bx$-update, we use the early termination technique described in \cite[\S 4.3.2]{Boyd:2011:DOS:2185815.2185816}. This is accomplished by splitting the ADMM algorithm into two parts: first the algorithm is run with $\stopthreshprime=1$, $\stopthreshdual = 10^{-4}$, $\gradstopthresh=10^{-3}$. Then, the thresholds are set to the final values of $\stopthreshprime=0.5$, $\stopthreshdual = 10^{-6}$, $\gradstopthresh=10^{-8}$.

While we used gradient descent for its simplicity and robustness, a possible enhancement would be to use an optimization algorithm with faster convergence.

\section{Coherence Derivations} \label{Sec:CoherenceDerivations}

The quasi-homogeneous approximation is
\begin{align} \label{eqn:quasihomogeneous}
\Coh(\br, \brho)
&= \Intensity(\br) \exp\left( -\frac{ \norm{\brho}_2^2 }{2 \cohwidth^2} \right),
\end{align}
where $\cohwidth$ is termed the coherence width.
In this approximation, the function is separable with regard to the ``intensity'' and ``coherence'' components \cite{Carter:77,Baleine:04}.

Under the Fresnel approximation, the impulse response function for the electric field in free space is
\begin{equation}
h(\br)=i \frac{\exp(-ikd)}{\lambda d}\exp\left[ - \frac{i k \norm{\br}_2^2}{2 d}\right].
\end{equation}
Then, the propagation is given by
\begin{align} \label{eqn:coh_prop_unrotated}
\CohUnrotated_d(\br_1, \br_2)
&= \frac{1}{(\lambda d)^2} \iint_{\bbR^2 \times \bbR^2} 
\,d^2\br_1' \,d^2\br_2'
\, \CohUnrotated(\br_1',\br_2')
\nonumber \\
&\quad \times h(\br_1'-\br_1) \, h^*(\br_2'-\br_2)
\nonumber \\
&= \frac{1}{(\lambda d)^2} \iint_{\bbR^2 \times \bbR^2} \,d^2\br_1' \,d^2\br_2'
\, \CohUnrotated(\br_1', \br_2')
\nonumber \\
&\quad \times \exp \left[ \frac{i k}{d} \left( \norm{\br_1-\br_1'}_2^2 - \norm{\br_2-\br_2'}_2^2 \right) \right]
\end{align}
Applying the transformation $\br_1,\br_2 \rightarrow \br,\brho$, with $\br = \frac{\br_1+\br_2}{2}$ (the center point) and $\brho = \br_1-\br_2$ (the separation), we get
\begin{align} \label{eqn:Propagation_Integral2}
\Coh_d(\br, \brho)
&= \frac{1}{(\lambda d)^2} \iint_{\bbR^2 \times \bbR^2} \,d^2\br' \,d^2\brho' \, \Coh(\br', \brho')
\nonumber \\
& \qquad \times \exp \left[ \frac{i k}{d} (\br-\br') \dotprod (\brho-\brho') \right],
\end{align}
After substituting \eqref{eqn:quasihomogeneous} into \eqref{eqn:Propagation_Integral2}, and rearranging the integrals,
\begin{align} \label{eqn:PropagationIntegralCohSTEP1}
&\Coh_d(\br, \brho)
= \frac{\exp \left( \frac{i k}{d} \br \dotprod \brho \right) }{(\lambda d)^2}
\int_{\bbR^2} \,d^2\br' \, \exp \left( - \frac{i k}{d} \brho \dotprod \br' \right) \sourceimg(\br')
\nonumber \\
&\times \int_{\bbR^2} \,d^2\brho' \, \exp \left( - \frac{i k}{d} (\br - \br') \dotprod \brho' \right) \exp\left( -\frac{ \norm{\brho'}_2^2 }{2 \cohwidth^2} \right)
\end{align}
Then, evaluating the inner integral yields \eqref{eqn:PropagationIntegralCoh}.

We represent the transfer function of the wall as a Bidirectional Reflectance Distribution Function (BRDF) \cite{Nicodemus:65}
\begin{align}
h(\br, \btheta - \btheta')
&= \exp\left( -\frac{(\theta_x - \theta_x')^2}{2 \wallsd_x^2} - \frac{(\theta_y - \theta_y')^2}{2 \wallsd_y^2} \right).
\end{align}
Here, $\btheta = (\theta_x,\theta_y)$ is the angle between the surface normal and the incident vector along the $x$ and $y$ axes, while $\btheta'$ is the reflected angle defined in a similar way.  

The link between coherence and the angular spread of the light (in this context referred to as specific intensity) is via the relation \cite{Pierrat:05,Shen:17}
\begin{align} \label{eqn:CohSI_relation}
\SI_d(\br, \bu_{\perp})
&= \left( \frac{k}{2 \pi} \right)^2 |u_z|
\int_{\bbR^2} \,d^2 \brho
\nonumber \\
&\qquad \times \Coh_d(\br, \bu) \exp(-i k \brho \bu_{\perp}).
\end{align}
where $u_z$ is the $z$ component of unit length vector $\bu = (\bu_{\perp}, u_z)$.
Given that $\btheta-\btheta'$ is small due to the paraxial approximation and narrow spread of the specular reflection \cite{TAVASSOLY2004252}, then $\bu_{\perp} - \bu_{\perp}' \approx \btheta - \btheta'$ and we can can calculate the scattered \textit{specific intensity} using the convolution
\begin{align} \label{eqn:scatteredSI}
\SI_S(\br, \bu_{\perp})
\approx \int_{\bbR^2} \,d^2 \bu_{\perp}' \SI_d(\br, \bu_{\perp}') \hat{h}(\br, \bu_{\perp}-\bu_{\perp}' )
\end{align}
Calculating the scattered \textit{coherence} from \eqref{eqn:scatteredSI} using the inverse of \eqref{eqn:CohSI_relation}
yields \eqref{eqn:scatteredCoh}.
Substituting $\brho = \bzero$ in \eqref{eqn:scatteredCoh} gives
\begin{align} \label{eqn:PropagationIntegralInt2}
\Intensity_d(\br)
&= \frac{2 \pi \cohwidth^2}{\lambda^2 d^2}
\int_{\bbR^2} \,d^2\br' \, 
\sourceimg(\br')
\exp\left( -\frac{\cohwidth^2 k^2 \norm{ \br - \br' }_2^2}{2 d^2} \right),
\nonumber
\end{align}
and hence \eqref{eqn:PropagationIntegralInt}.

While we use a single wavelength for simplicity, the propagation of broadband light can also be accomplished by propagating at multiple wavelengths and summing the results.  This method would still preserve the linearity of the transforms.
In fact, exploiting the multiple wavelengths could improve image quality, although this is beyond our scope here.

\section*{Acknowledgment}
The authors thank A. Abouraddy and A. Dogariu of the College of Optics and Photonics, University of Central Florida, for very useful discussions concerning light propagation in scattering media.

\ifCLASSOPTIONcaptionsoff
  \newpage
\fi

\bibliographystyle{IEEEtran}
\bibliography{main}

\end{document}